%% file: iclr2026_conference.tex
\documentclass{article} % For LaTeX2e
\usepackage{iclr2026_conference,times}

\input{math_commands.tex}

\usepackage{hyperref}
\usepackage{url}
\usepackage{enumerate}
\usepackage[T1]{fontenc}    % use 8-bit T1 fonts
\usepackage{booktabs}       % professional-quality tables
\usepackage{amsfonts}       % blackboard math symbols
\usepackage{nicefrac}       % compact symbols for 1/2, etc.
\usepackage{microtype}      % microtypography
\usepackage{xcolor}         % colors
\usepackage{graphicx} 
\usepackage{multirow}
\usepackage{amsmath}
\usepackage{cleveref} % for clever cross-referencing
\usepackage{subcaption}
\usepackage{setspace}
\usepackage{enumitem}
\usepackage{array}
\usepackage{xspace}
\usepackage{pifont}
\usepackage{paralist}
\usepackage[normalem]{ulem}
\usepackage{algorithm}
\usepackage{algorithmic}
\usepackage{wrapfig}
\usepackage{bbding}
\usepackage{adjustbox}
\usepackage{amsthm}
\usepackage{amssymb}
\usepackage{caption}
\usepackage{subcaption}
\usepackage{tabularx}
\usepackage{colortbl}
\usepackage{algorithm}
\usepackage{algorithmic}
\usepackage{makecell}
\usepackage{titlesec}
\titlespacing*{\paragraph}{0pt}{0.2ex plus 0.5ex minus 0.2ex}{0.5em}

\definecolor{color1}{HTML}{D0E2BF} 
\definecolor{color2}{HTML}{F7E6D8}
\definecolor{seagreen}{HTML}{2E8B57}
\definecolor{firebrick}{HTML}{B22222}
\definecolor{tabblue}{HTML}{5b9ad5}
\definecolor{taborange}{HTML}{ed7d31}
\newcommand{\our}{LATR}
\newcommand{\old}{Stochastic Sampling}
\newcommand{\imp}[1]{\textcolor{seagreen}{(#1)}}
\newcommand{\drp}[1]{\textcolor{firebrick}{(#1)}}

\newcommand{\modify}[1]{{#1}}

\title{Lookahead Tree-Based Rollouts for Enhanced Trajectory-Level Exploration in Reinforcement Learning with Verifiable Rewards}

\author{
Shangyu Xing$^1$ 
\quad
Siyuan Wang$^2$\thanks{ Correspondence to \url{sw_641@usc.edu} and \url{xiangren@usc.edu}.}
\quad
Chenyuan Yang$^3$
\quad
Xinyu Dai$^1$
\quad
Xiang Ren$^2\footnotemark[1]$
\\
$^1$ Nanjing University \quad $^2$ University of Southern California \quad $^3$ Fudan University
}
\iclrfinalcopy

\begin{document}
\maketitle

\begin{abstract}
Reinforcement Learning with Verifiable Rewards (RLVR), particularly with algorithms like Group Relative Policy Optimization (GRPO), has proven highly effective in enhancing the reasoning capabilities of large language models. However, a critical bottleneck in current pipelines lies in the limited diversity of sampled trajectories during group rollouts. Homogeneous trajectories and their associated rewards would diminish the return signals for policy updates, thereby hindering effective policy learning. This lack of diversity stems primarily from token-level stochastic sampling, where local variations are likely to collapse into near-identical reasoning paths. To address this limitation, we propose Lookahead Tree-Based Rollouts (\our), a novel rollout strategy designed to explicitly promotes trajectory-level diversity by enforcing branching into different candidate tokens likely to yield distinct continuations.  
Specifically, \our\ iteratively operates in three stages: (1) branching at high-uncertainty generation steps, (2) performing lookahead simulation for each new branch, and (3) pruning branches that exhibits prolonged similarity during simulation. 
% By applying these stages, \our\ enhances trajectory-level exploration and produces more diverse rollout groups. 
Compared with \old, \our\ accelerates policy learning by 131\% and improves final pass@1 performance by 4.2\% on both GRPO and Dynamic sAmpling Policy Optimization (DAPO) algorithms across different reasoning tasks. Our code and data are available at \url{https://github.com/starreeze/latr}.
\end{abstract}

\input{Sections/intro}

\input{Sections/methodology}

\input{Sections/experiments}

\input{Sections/discussion}

\input{Sections/related_work}

\section{Conclusion}
In this work, we present Lookahead Tree-Based Rollout, a novel rollout strategy that explicitly promotes trajectory-level diversity in RLVR by dynamically branching at high-uncertainty tokens and pruning non-divergent paths via lookahead simulation. By moving beyond token-level sampling heuristics, \our\ enriches policy learning signals, accelerating training convergence while improving final performance by a large margin across different benchmarks. Our work demonstrates that trajectory-level rollout diversity is key to scaling RLVR effectively and efficiently.

\section*{Reproducibility Statement}
To better support reproducibility, we explain all the details to reproduce our results in Section \ref{sec:detail}, including the parameters for our methods, training details, environment and framework versions.


\section*{Acknowledgments}
This research is supported in part by the Office of the Director of National Intelligence (ODNI), Intelligence Advanced Research Projects Activity (IARPA), via the HIATUS Program contract \#2022-22072200006, the Defense Advanced Research Projects Agency with award HR00112220046, and NSF IIS 2048211. We would like to thank all the collaborators in USC INK research lab for their constructive feedback on the work.

\bibliography{iclr2026_conference}
\bibliographystyle{iclr2026_conference}

\input{Sections/appendix}

\end{document}

%% file: math_commands.tex
\usepackage{amsmath,amsfonts,bm}

\def\eqref#1{equation~\ref{#1}}
\def\1{\bm{1}}

\DeclareMathAlphabet{\mathsfit}{\encodingdefault}{\sfdefault}{m}{sl}
\SetMathAlphabet{\mathsfit}{bold}{\encodingdefault}{\sfdefault}{bx}{n}

\DeclareMathOperator*{\argmax}{arg\,max}

%% file: Sections/intro.tex
\begin{figure*}[h]
    \centering
    \includegraphics[width=\linewidth]{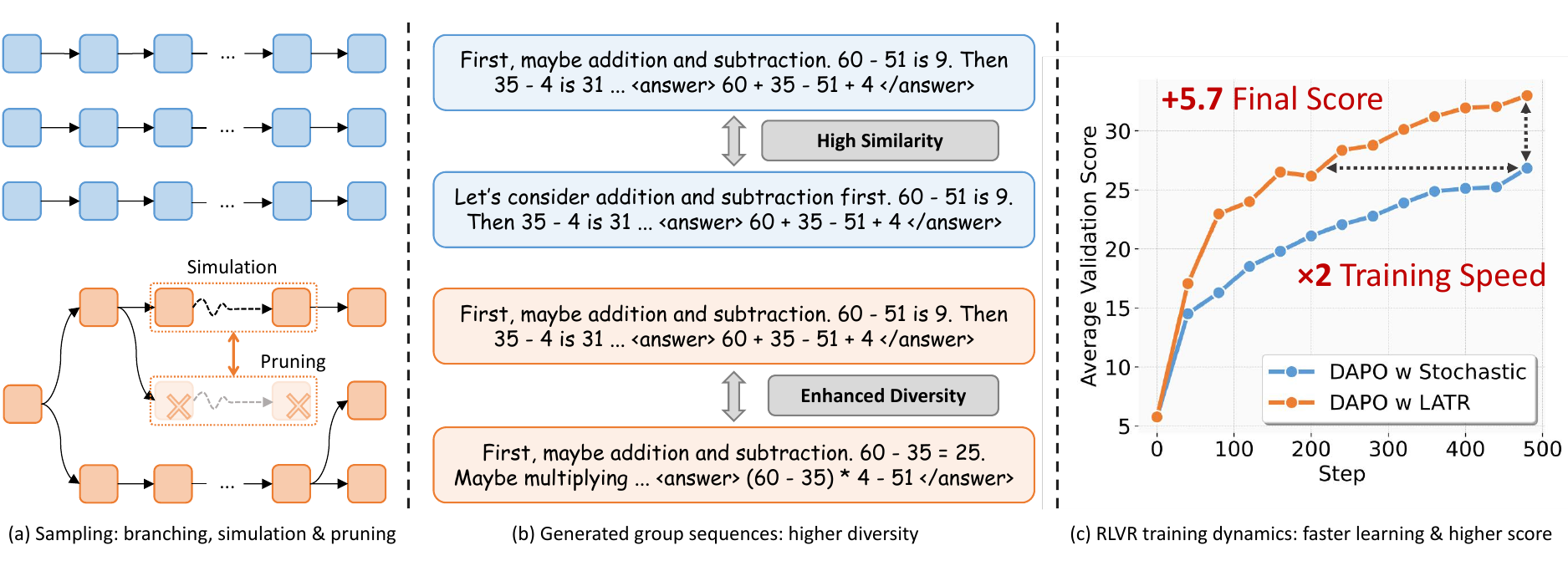}
    \caption{Comparison of conventional token-level \textbf{\textcolor{tabblue}{stochastic sampling}} and our proposed method \textbf{\textcolor{taborange}{\our}} on sampling process, rollout sequence diversity, and performance on DAPO Math dataset.}
    \label{fig:overview}
\end{figure*}

\section{Introduction}

Reinforcement learning with verifiable rewards (RLVR) has emerged as a powerful paradigm for enhancing the reasoning capabilities of large language models recently ~\citep{r1, qwen3, gpt5}. By leveraging sequence rollouts and updating policies according to appropriate rewards, RLVR can significantly improve performance across diverse reasoning tasks, including mathematical problem solving, code generation, and multi-step logical deduction~\citep{tinyzero}. Algorithms such as Group Relative Policy Optimization (GRPO)~\citep{deepseekmath} have become central to this approach, enabling stable model training through in-group trajectory comparisons to learn from high-quality responses while penalizing low-rewarded ones. 

A key challenge in these methods lies in the limited diversity of trajectories sampled during the rollout phase~\citep{8020, negative}. When trajectories within a group exhibit high similarity, the estimated relative advantage and learning signals tend to diminish. As a result, the policy updates become less informative, ultimately hindering the effective scaling.
Recent efforts have sought to mitigate this issue through various approaches, including increasing sampling temperature~\citep{prorl} and dynamically filtering out groups with highly similar samples~\citep{dapo}. However, the former focuses on token-level variation without ensuring trajectory-level divergence, while the latter relies on post hoc filtering that provides only limited within-group diversity at the cost of excessive over-generation. Both methods therefore yield only modest improvements in rollout diversity under a constrained generation budget.
% , add regularization to objective function~\citep{yu2025dapo}, training on high-entropy tokens~\citep{80/20}, and using pass@K reward~\citep{pass@k Training} to enhance the exploration ability of LLMs.
% indirect regularization techniques. 
% For example, DAPO~\citep{} proposes to increase the upper bound of the importance sampling ratio to leave more room for the increase of low-probability tokens and dynamic sampling to skip prompts with identical rewards. Pass@k Training~\citep{} replaces the standard pass@1 reward with pass@k metric to encourage the model to achieve higher pass@k performance, thereby enhancing its exploration ability.
% owes its superior performance majorly to its group filtering technique, where over-sampling is enforced and rollout groups with identical rewards are removed. Although these methods represent meaningful progress, they remain reactive rather than proactive: they adjust rollout groups post hoc but do not explicitly structure the exploration process or prune redundant paths during generation. As a result, they often fail to prevent the emergence of semantically overlapping trajectories, particularly under tight computational constraints where every sampled sequence must contribute meaningfully to policy learning.

We argue that such diversity limitation stems from the predominate reliance on token-level stochastic sampling strategies, where each sequence in a group is generated independently by sampling tokens from the model's output distribution at each decoding step. While simple and widely adopted, this approach ignores the contrast among sequences within the group and fails to enforce distinction or complementarity at the trajectory level, thus exhibiting an inherently myopic limitation.
Specifically, token-level variations typically occur without lookahead ability, making local deviations (e.g., substituting ``compute'' with ``calculate'') easily collapse back into nearly identical reasoning paths, leading to redundant exploration and diminishing returns.
% it operates myopically at the token level and lacks awareness of higher-order trajectory structure. Consequently, even when local token variations occur, e.g., substituting ``compute'' with ``calculate'', the underlying reasoning path often remains functionally identical, leading to redundant exploration and inefficient use of the rollout budget.

% (Highlight the importance of exploration diversity during sequence rollout ... (citing different variants of GRPO such as DAPO / GSPO that also optimize in terms of exploration). But they are limited in using an indirect mechanism to improving diversity by ... and still rely on token-level stochastic sampling ...)
% However, the current implementation relies on token-level stochastic sampling, which is prone to inefficient and limited exploration. Specifically, at each step, tokens are sampled independently from the actor model’s output distribution, treating each sequence in isolation within the group. This probability-driven sampling strategy neglects higher-level structural and semantic differences across sequences, often resulting in similar reasoning paths despite local token variations. As a consequence, the rollout group tends to exhibits reduced diversity, especially under constrained rollout budgets, and ultimately leading to suboptimal policy updates. 

To address these limitations, we propose \textbf{L}ook\textbf{a}head \textbf{T}ree-Based \textbf{R}ollout (\our), a strategy designed to explicitly promote trajectory-level diversity within a group by maintaining rollouts in a tree structure.
At token positions with high model uncertainty, \our\ enforces branching into different candidate tokens that are highly likely to yield distinct continuations.
To guarantee that each selected candidate token can lead to a different reasoning path, \our\ performs lookahead simulation by continuing generation for a fixed length, and removes those candidates failing to diverge from others. This branching, simulation and pruning procedure is iteratively repeated until the target number of rollouts is reached, after which all surviving partial sequences continue to be extended in parallel under standard stochastic sampling.
This ensures that the generated trajectories are reasonably distinct from each other, thereby enriching the in-group rollout diversity.
% Unlike independent stochastic sampling, \our\ maintains a dynamically expanding and pruned tree of reasoning trajectories. Two interleaved mechanisms drive this process: \emph{branching}, which spawns new branches at points of high model uncertainty by simultaneously exploring multiple trajectories; and \emph{pruning}, which discards branches that fail to meaningfully diverge from others over successive steps.  
% a novel rollout strategy that replaces independent token sampling with a dynamically maintained and pruned tree search over reasoning trajectories. Rather than treating each sequence in isolation, \our\ jointly explores multiple reasoning paths
% through two interleaved mechanisms: Branching, which spawns new trajectories simultaneously with multiple candidate tokens at points of high model uncertainty; Pruning, which deletes branches that fail to meaningfully diverge from others in prolonged steps.
% by branching at points of high model uncertainty and subsequently pruning branches that fail to meaningfully diverge. 

We apply \our\ strategy to both GRPO and DAPO algorithms and evaluate across 5 datasets involving mathematical and logical reasoning. Our experiments demonstrate that \our\ consistently accelerates policy learning by an average of 131\%, while simultaneously improving final task performance of pass@1 by averagely 4.2\% across different tasks. Our contributions are summarized as follows:
\begin{compactenum}[1)]
    \item We introduce a novel tree-based rollout algorithm \our\ that explicitly optimizes for trajectory-level diversity, which can be integrated seamlessly into any policy update algorithms.
    \item We provide extensive empirical validation across tasks and training configurations, demonstrating consistent improvements over existing sampling strategies in RLVR pipelines.
\end{compactenum}

%% file: Sections/methodology.tex
\begin{figure*}[t]
    \centering
    \includegraphics[width=\linewidth]{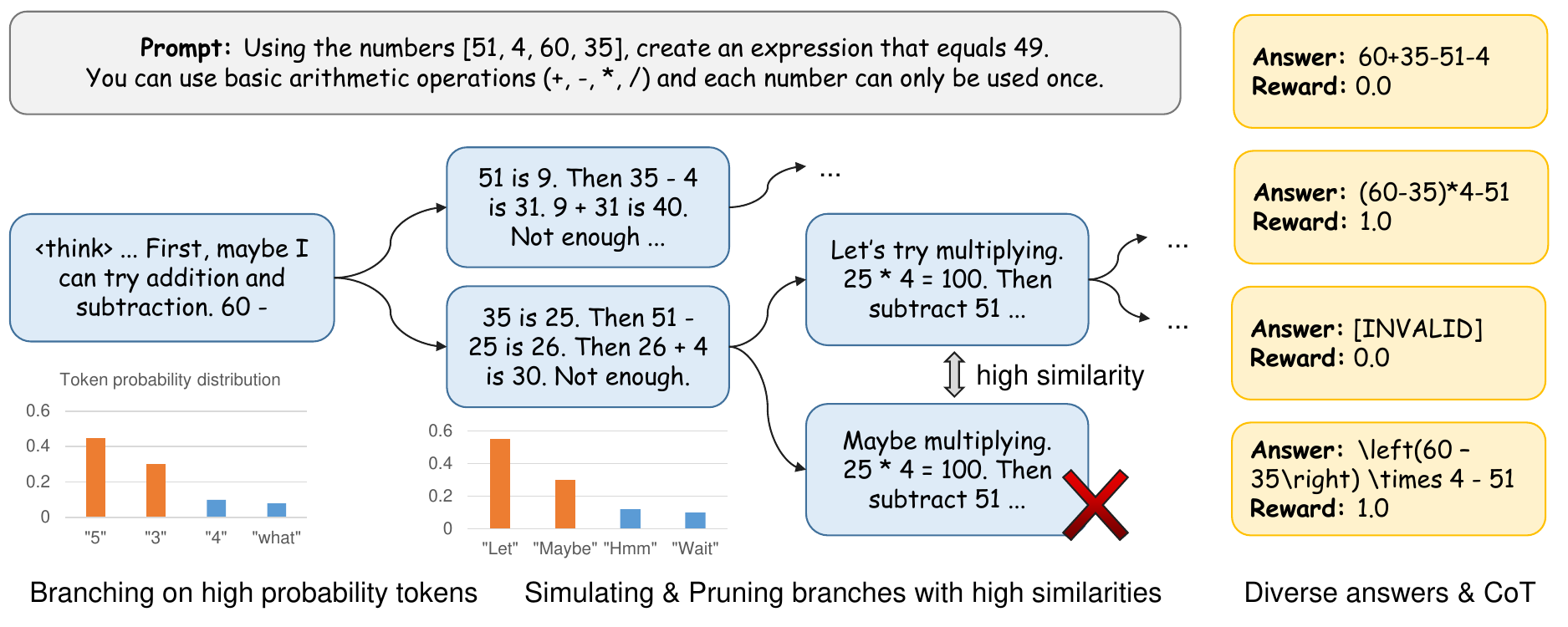}
    \caption{An overview of \our. A dynamic search tree is built by branching on model uncertainty, simulating and pruning similar branches, resulting in diverse answers and reasoning paths.}
    \label{fig:method}
\end{figure*}

\section{Preliminary}
We adopt Group Relative Policy Optimization (GRPO) \citep{deepseekmath} as the foundational RL algorithm for policy refinement. Unlike Proximal Policy Optimization (PPO) \citep{ppo}, which relies on a learned value function to estimate advantages, GRPO eschews explicit value modeling and instead computes advantages directly from group-relative rewards. This design simplifies training dynamics and enhances stability in reward-sparse or high-variance environments. 

Each training iteration in GRPO consists of two phases: (1) \textit{rollout}, where multiple candidate responses are sampled per prompt, and (2) \textit{policy update}, where the policy is optimized using group-normalized advantages and a clipped surrogate objective.

\paragraph{Rollout.}
Given a prompt \( p \) drawn from the dataset \( \mathcal{D} \), a group of \( k \) candidate sequences \( \{s_i\}_{i=1}^k \) are generated via autoregressive sampling from policy \( \pi_\theta \). Each sequence is constructed token-by-token through stochastic sampling from the model's predicted next-token distribution. This process is identical to inference-time generation.

Formally, let \( S_l \) denote the multiset of partial sequences of length \( l \) generated for prompt \( p \). The rollout process is recursively defined as:
\begin{equation}
    S_0 = \{ \underbrace{\epsilon, \epsilon, \dots, \epsilon}_k \}, \quad
    S_{l+1} = \bigcup_{s \in S_l} \left\{ s \oplus t \mid t \sim \pi_\theta(\cdot \mid p \oplus s) \right\},
\end{equation}
where \( \epsilon \) is the empty sequence, \( \oplus \) denotes token concatenation, and sampling terminates when all sequences reach an end-of-sequence token or a maximum length $n$. The final output is the group \( S =  \{s_1, \dots, s_k\} \). This group-based sampling enables direct comparison of responses under the same prompt, forming the basis for relative advantage estimation.

\paragraph{Policy Update.}
Policy update aims to refine policy $\pi$ by maximizing expected cumulative rewards. Similar to PPO, GRPO adopts a clipped objective, together with a directly imposed KL penalty term:
\begin{multline}
    \mathcal{J}_{\text{GRPO}}(\theta) = \mathbb{E}_{p\sim\mathcal{D}, \{s_i\}_{i=1}^k \sim \pi_{\theta_{\text{old}}}(\cdot|p)} \\
    \left[ \frac{1}{k} \sum_{i=1}^k \frac{1}{|s_i|} \sum_{l=1}^{|s_i|} \left( \min\left(r_{i,l}(\theta)\hat{A}_{i,l},\ \text{clip}\left(r_{i,l}(\theta), 1-\varepsilon, 1+\varepsilon\right)\hat{A}_{i,l} \right) - \beta D_{\text{KL}}(\pi_\theta || \pi_{\text{ref}}) \right) \right],
\end{multline}
where the advantage $\hat{A}$ is calculated by normalizing the group-level rewards $\{R_i\}_{i=1}^k$, and the ratio $r$ compares the likelihood of token $s_{i,l}$ under the current and old policies:
\begin{equation}
    \hat{A}_{i,l} = \frac{R_i - \text{mean}(\{R_i\}_{i=1}^k)}{\text{std}(\{R_i\}_{i=1}^k)}, \quad
    r_{i,l}(\theta) = \frac{\pi_\theta(s_{i,l} \mid p, s_{i,<l})}{\pi_{\theta_{\text{old}}}(s_{i,l} \mid p, s_{i,<l})}.
\end{equation}

\paragraph{Variations of GRPO.}
Building upon this, DAPO \citep{dapo} improves GRPO in several aspects. In rollout stage, DAPO oversamples data batches and filters out groups with identical rewards. If the retained groups are insufficient to fill a batch, additional rollouts are iteratively sampled. This mechanism trades computational efficiency for higher response diversity and more informative gradients. In policy update stage, DAPO addresses GRPO's limitations in long-form generation tasks by implementing token-level loss calculation to mitigate length bias, and employs decoupled clipping without RL penalty to encourage exploration. Formally, the objective is
\begin{align}
    \mathcal{J}_{\text{DAPO}}(\theta) = ~~& \mathbb{E}_{p\sim\mathcal{D}, \{s_i\}_{i=1}^k \sim \pi_{\theta_{\text{old}}}(\cdot|p)} \notag \\&
    \left[ \frac{1}{\sum_{i=1}^k |s_i|} \sum_{i=1}^k \sum_{l=1}^{|s_i|} \min\left(r_{i,l}(\theta)\hat{A}_{i,l},\ \text{clip}\left(r_{i,l}(\theta), 1-\varepsilon_\text{low}, 1+\varepsilon_\text{high} \right)\hat{A}_{i,l} \right) \right].
\end{align}
These enhancements make DAPO a more robust and effective algorithm for complex reasoning tasks.

\section{Lookahead Tree-Based Rollout}

\begin{wrapfigure}{r}{0.53\textwidth}
\vspace{-2em}
\hspace{0.4em}
\begin{minipage}{0.5\textwidth}
\begin{algorithm}[H]
\caption{Lookahead Tree-Based Rollouts}
\label{alg:sample}
\begin{algorithmic}[1]
\REQUIRE Policy model $\pi$, rollout number $k$, \\
prompt $p$, absolute branching threshold $\tau_{\text{abs}}$, \\
relative threshold $\tau_{\text{rel}}$, pruning threshold $\tau_{\text{ed}}$, \\
lookahead step $r$, max length $n$.
\ENSURE Set of rollouts $S = \{s_1, \dots, s_k\}$.
\STATE Initialize $S \gets \{\epsilon\}$ \hfill $\triangleright$ Single root branch
\FOR{$l = 1$ to $n$}
    \STATE $S_{\text{next}} \gets \emptyset$
    \STATE
    \STATE \COMMENT{- - - - - - Branching logic - - - - - -}
    \FOR{branch $s_i \in S$}
        \STATE $\mathcal{P}_i \gets \pi(\cdot~|~p \oplus s_i)$ \hfill $\triangleright$ Prob distribution
        \STATE $c_i^\star \gets \argmax_c \mathcal{P}_i[c]$ \hfill $\triangleright$ Top candidate
        \STATE $s_i^{\text{extend}} \gets s_i \oplus c_i^\star$ \hfill $\triangleright$ Extend main
        \STATE $S_{\text{next}} \gets S_{\text{next}} \cup \{s_i^{\text{extend}}\}$
        \STATE $\mathcal{C}_i \gets \{c \neq c_i^\star \mid \mathcal{P}_i[c] > \tau_{\text{abs}}$ and \\ \hspace{2.7em}
        $\mathcal{P}_i[c_i^\star] - \mathcal{P}_i[c] < \tau_{\text{rel}}\}$
        
        \FOR{$c \in \mathcal{C}_i$}
            \IF{$|S_{\text{next}}| < k$}
                \STATE $s_{\text{new}} \gets s_i \oplus c$ \hfill $\triangleright$ New branch
                \STATE $s_{\text{new}}.\text{parent} \gets s_i$
                \STATE $s_{\text{new}}.\text{birth} \gets l$
                \STATE $S_{\text{next}} \gets S_{\text{next}} \cup \{s_{\text{new}}\}$
            \ENDIF
        \ENDFOR
    \ENDFOR
    \STATE
    \STATE \COMMENT{- - - - - - Pruning logic - - - - - -}
    \FOR{$s_j \in S_{\text{next}}$ with $s_j.\text{birth} = l - r$}
        \IF{$\text{EditDist}(s_j, s_j.\text{parent}) < \tau_{\text{ed}}$}
            \STATE Remove $s_j$ with its descendants
        \ENDIF
    \ENDFOR
    \STATE
    \STATE $S \gets S_{\text{next}}$
    
\ENDFOR
\RETURN Pad $S$ to exactly $k$ sequences
\end{algorithmic}
\end{algorithm}
\end{minipage}
\vspace{-3em}
\end{wrapfigure}

To address the limited diversity of conventional token-level sampling during the rollout phase, we introduce \textbf{L}ook\textbf{a}head \textbf{T}ree-Based \textbf{R}ollout (\our), a structured exploration strategy inspired by Monte Carlo Tree Search \citep{mcts}. \our\ achieves diverse trajectory generation by enforcing branching at candidate tokens that are highly likely to yield distinct continuations. 

Specifically, \our\ operates through three iterative stages: (1) \textit{Branching}, which creates new trajectories at token positions with high model uncertainty; (2) \textit{Lookahead Simulation}, where the new branch is extended for a fixed lookahead window of $r$ tokens; and (3) \textit{Pruning}, where simulated sequences that fail to diverge from others are removed. This process repeats until the target number of rollouts is reached, ensuring their diversity. We provide the complete algorithm in Algorithm \ref{alg:sample} and an illustration in Figure \ref{fig:method}.

\subsection{Branching}

\our\ begins with a root node corresponding to the input prompt. At each generation step $l$, every active branch is extended by its highest-probability token to ensure progress along the most likely trajectory. These branches are regraded as parent branches. Simultaneously, if other candidate tokens satisfy both the absolute probability threshold $\tau_{\text{abs}}$ and the relative probability threshold $\tau_{\text{rel}}$, new child branches are instantiated. This dual-threshold mechanism targets reasoning crossroads where the model is genuinely uncertain between semantically distinct continuations, while preventing the branches diverging too far from the policy distribution. Branching allows \our\ to maintain multiple distinct reasoning paths in parallel, significantly increasing the probability of discovering high-quality, diverse solutions.

Formally, let $S_l$ denote the set of active branches at step $l$, and for each branch $s \in S_l$, let $\mathcal{P}_s$ denote its next-token distribution, the most likely token $c_s^\star = \argmax_c \mathcal{P}_s[c]$, and $\mathcal{C}_s$ is the set of all remaining candidates excluding $c_s^\star$. A new child branch $s \oplus c$ is created if:
\begin{equation}
\label{eq:branch}
    \mathcal{P}_s[c] > \tau_{\text{abs}} \quad \text{and} \quad \mathcal{P}_s[c_s^\star] - \mathcal{P}_s[c] < \tau_{\text{rel}}.
\end{equation}
The set of branches after expansion is the union of the parent branches with their new children:
\begin{equation}
    S_l' = \bigcup_{s \in S_l} \left( \{ s \oplus c_s^\star \} \cup \left\{ s \oplus c \mid c \in \mathcal{C}_s,~ \text{conditions of (\ref{eq:branch}) hold},~ |S_l'| < k \right\} \right).
\end{equation}
If the rollout budget $k$ is reached, candidate branches are prioritized by descending probability $\mathcal{P}_s[c]$. This ensures that more plausible alternatives are more likely to be explored.

\subsection{Simulation \& Pruning}

While the above branching strategy effectively enables structured parallel exploration, it faces two challenges: (1) unconstrained branching leads to exponential growth, quickly exhausting the rollout budget and limiting exploration sequentially; (2) branches started from token-level variations easily collapse back into nearly identical reasoning paths, struggling to ensure trajectory-level diversity.
% many branches may represent superficial variations that ultimately converge to identical reasoning, impacting the diversity of final sequences.

To address these issues, \our\ incorporates a lookahead simulation and pruning phase. After branching, each new trajectory continues generation for a fixed lookahead window of $r$ tokens. These continuations are then evaluated for divergence using normalized edit distance, and branches exhibiting insufficient divergence from their parents are pruned.

Specifically, at each step $l$, \our\ identifies all branches $s$ created at step $l-r$ and computes the normalized edit distance over their most recent $r$ tokens relative to their parents' corresponding segment. If the distance falls below a threshold $\tau_\text{ed}$, the branch and all its descendants are removed:
\begin{equation}
    \label{eq:prune}
    S_l^\text{prune} = \left\{ s ~\middle|~ s \in S_l',~ s.\text{birth} = l - r ,~ \text{EditDist}(s[-r:], s.\text{parent}[-r:]) < \tau_{\text{ed}} \right\},
\end{equation}
\begin{equation}
    S_{l+1} = \left\{ s ~\middle|~ s \in S_l',~ s \notin S_l^\text{prune} \right\},
\end{equation}
where EditDist() indicates normalized edit distance, i.e., the Levenshtein distance between token ID sequences, divided by sequence length. This ensures that only branches exhibiting meaningful divergence within the lookahead window are preserved. We also explore similarity measures other than edit distance in Appendix \ref{app:similarity}, and find that their performance are very close. Notably, \our\ is backtracking-free, so the number of forward passes required by a group rollout is bounded by $O(nk)$, where $k$ is the rollout number (tree width) and $n$ is the maximum completion length (tree depth).

Through lookahead simulation and pruning, \our\ preserves only diverse branches that are more likely to yield distinct reasoning paths. The final output consists of $k$ surviving branches, padded if necessary to meet the rollout number requirements. The entire procedure is compatible with any autoregressive language model and can be integrated seamlessly into existing policy update algorithms and RLVR frameworks without modifications.

\subsection{Further Optimizations}
\label{sec:optim}

\paragraph{Early Stopping.}
When the tree width reaches the rollout number $k$, \our\ has already produced $k$ sequences that are likely to lead to diverse reasoning paths. At this point, the generation process is switched to standard stochastic sampling for all remaining steps. This allows surviving branches to continue exploring the solution space stochastically while maintaining the diversity benefits from \our. Analyses on the stopping length is in Appendix \ref{app:stat}.

\paragraph{Hybrid Rollout for RL Training.}
While \our\ excels at promoting diverse exploration during RL training, its explicit divergence objective can create a mismatch with test-time behavior. At real-world inference, models typically generate a single trajectory using greedy or stochastic decoding, prioritizing correctness and coherence over diversity. However, policy updates with \our\ tries to maximize the reward from the \our-generated diverse rollout group. Training exclusively with \our\ throughout the entire process may thus bias the policy toward over-exploration patterns that do not generalize. To bridge this gap, we adopt a hybrid sampling strategy during RL training. At each training step, we allocate a fraction $\eta$ of rollouts to \our\ and the remainder to standard \old:
\begin{equation}
    k_{\text{\our}} = \lfloor \eta k \rceil, \quad k_{\text{std}} = k - k_{\text{\our}},
\end{equation}
where $\lfloor \cdot \rceil$ denotes rounding to the nearest integer. We anneal $\eta$ exponentially over training step $i$:
\begin{equation}
    \eta = \eta_0 \cdot \gamma^i,
\end{equation}
with decay rate $\gamma < 1$. This ensures early-stage exploration benefits from \our's diversity, while later stages increasingly mimic test-time behavior to reduce train-test discrepancy.

%% file: Sections/experiments.tex
\section{Experiments}
\subsection{Experimental Setup}
% To rigorously evaluate our method in reasoning-intensive environments, we design experiments around two canonical tasks suited for RLVR: logical reasoning and mathematical problem solving. For logical reasoning, we use the Countdown dataset for training and evaluation; for mathematical problem solving, we train the model on DAPO Math dataset and evaluate it on additional datasets, including MATH500 (?), AMC2023 (?), and OlympiadBench (?). Consistent with \citet{tinyzero, dapo}, we design the reward function for the Countdown task with a format reward and correctness reward, weighting 0.1 and 0.9. For math tasks, reward function is binary, assigning 1.0 to correct answers. For evaluation, we sample 8 times for each question, reporting correctness scores and average completion length of Pass@1 and Pass@8. For other details, including the introduction to datasets, explanation on evaluation protocol, and hyperparameter settings, please refer to Appendix \ref{app:detail}.

To rigorously assess \our’s performance in reasoning-intensive environments, we evaluate it on two canonical domains suited for RLVR: logical reasoning and mathematical problem solving.

\paragraph{Logical Reasoning.}  
We adopt the Countdown dataset for both training and evaluation. Following prior work \citep{tinyzero}, we use reward $R = 0.1 \cdot R_{\text{format}} + 0.9 \cdot R_{\text{correctness}}$, where \(R_{\text{format}}\) encourages outputs with proper form and \(R_{\text{correctness}}\) assigns full reward for logically correct solutions.

\paragraph{Mathematical Problem Solving.}  
Models are trained on the DAPO-Math dataset and evaluated on three additional benchmarks: MATH-500 \citep{math500}, AMC-2023 \citep{amc}, and Olympiad-Bench \citep{olympiad}. Consistent with \citet{dapo}, the reward is binary: \(R = 1.0\) for correct final answers, and \(0\) otherwise.

\paragraph{Evaluation Protocol.}  
For each test instance, we sample 8 independent completions. We report Pass@1 and Pass@8 correctness scores along with the average completion length to assess solution conciseness. All implementation details, including dataset descriptions, hyperparameters, and environment configurations, are provided in Appendix \ref{app:detail}.

\subsection{Terminating Performance}

% \begin{wraptable}{r}{0.62\textwidth}
% \vspace{-1em}
% \hspace{0.4em}
% \begin{minipage}{0.6\textwidth}
\begin{table}[t]
    \centering
    \caption{Performance comparison of test correctness and average completion length on the \textbf{Countdown} dataset. $\uparrow$ indicates higher is better, while $\downarrow$ indicates lower is better. Relative improvement of \our\ to \old\ with the same policy update algorithm is marked in the parentheses, where \textcolor{seagreen}{green} indicates positive improvements and \textcolor{firebrick}{red} otherwise. Best results are in bold.}
    \adjustbox{width=0.7\textwidth}{
    \begin{tabular}{lcccc}
        \toprule
        \multirow{2}*{Method} & \multicolumn{2}{c}{Correctness (\%) $\uparrow$} & \multicolumn{2}{c}{Average Length $\downarrow$} \\
        & Pass@1 & Pass@8 & Pass@1 & Pass@8 \\
        \midrule
        Qwen2.5-3B                & 1.1 & 5.5 & 543 & 975   \\
        + GRPO w Stochastic & 65.9 & 73.9 & 473 & 610 \\
        + DAPO w Stochastic & 70.7 & 78.0 & 483 & 630 \\
        + GRPO w \our       & 70.9 \imp{+5.0} & 77.4 \imp{+3.5} & 378 \imp{-20\%} & 469 \imp{-23\%} \\
        + DAPO w \our       & \textbf{74.7} \imp{+4.0} & \textbf{81.5} \imp{+3.5} & \textbf{367} \imp{-24\%} & \textbf{453} \imp{-28\%} \\
        \bottomrule
    \end{tabular}}
    \label{tab:countdown}
\end{table}
% \end{minipage}
% \vspace{-1em}
% \end{wraptable}

\begin{table}[t]
    \centering
    \caption{Performance comparison on \textbf{DAPO Math} and \textbf{AMC 2023} dataset.}
    \adjustbox{width=\textwidth}{
    \begin{tabular}{lcccccccc}
        \toprule
        \multirow{3}*{Method} & \multicolumn{4}{c}{DAPO-Math (val)} & \multicolumn{4}{c}{AMC-2023} \\
        & \multicolumn{2}{c}{Correctness (\%) $\uparrow$} & \multicolumn{2}{c}{Average Length $\downarrow$} & \multicolumn{2}{c}{Correctness (\%) $\uparrow$} & \multicolumn{2}{c}{Average Length $\downarrow$} \\
        & Pass@1 & Pass@8 & Pass@1 & Pass@8 & Pass@1 & Pass@8 & Pass@1 & Pass@8 \\
        \midrule
        Qwen2.5-3B      & 5.6  & 20.1 & 938 & 2203 & 5.9 & 20.7 & 963 & 2255  \\
        + GRPO w Stoch. & 24.1 & 51.3 & 880 & 1732 & 32.8 & 59.7 & \textbf{833} & 1622 \\
        + DAPO w Stoch. & 26.8 & 53.1 & 1024 & 2022 & 37.8 & 62.7 & 1075 & 2116 \\
        + GRPO w \our   & 28.4 \imp{+4.3} & 51.9 \imp{+0.6} & \textbf{853} \imp{-3\%} & \textbf{1556} \imp{-10\%} & 35.6 \imp{+2.8} & 60.3 \imp{+0.6} & 838 \drp{+1\%} & \textbf{1537} \imp{-5\%} \\
        + DAPO w \our   & \textbf{32.5} \imp{+5.7} & \textbf{54.1} \imp{+2.8} & 896 \imp{-13\%} & 1880 \imp{-7\%} & \textbf{45.3} \imp{+7.5} & \textbf{65.0} \imp{+2.3} & 883 \imp{-18\%} & 1920 \imp{-9\%} \\
        \bottomrule
    \end{tabular}}
    \label{tab:math-1}
    \vspace{-0.5em}
\end{table}

\begin{table}[t]
    \centering
    \caption{Performance comparison on \textbf{MATH-500} and \textbf{Olympiad-Bench} dataset.}
    \adjustbox{width=\textwidth}{
    \begin{tabular}{lcccccccc}
        \toprule
        \multirow{3}*{Method} & \multicolumn{4}{c}{MATH-500} & \multicolumn{4}{c}{Olympiad-Bench} \\
        & \multicolumn{2}{c}{Correctness (\%) $\uparrow$} & \multicolumn{2}{c}{Average Length $\downarrow$} & \multicolumn{2}{c}{Correctness (\%) $\uparrow$} & \multicolumn{2}{c}{Average Length $\downarrow$} \\
        & Pass@1 & Pass@8 & Pass@1 & Pass@8 & Pass@1 & Pass@8 & Pass@1 & Pass@8 \\
        \midrule
        Qwen2.5-3B      & 24.7 & 54.4 & 748 & 1690 & 8.5 & 25.8 & 1088 & 2413   \\
        + GRPO w Stoch. & 58.4 & 76.7 & 657 & 1207 & 27.2 & 47.4 & 1058 & 2014 \\
        + DAPO w Stoch. & 60.4 & \textbf{79.2} & 700 & 1283 & 28.1 & 47.0 & 1162 & 2193 \\
        + GRPO w \our   & 61.9 \imp{+3.5} & 77.5 \imp{+0.8} & \textbf{594} \imp{-10\%} & \textbf{952} \imp{-25\%} & 29.5 \imp{+2.3} & \textbf{48.2} \imp{+0.8} & \textbf{954} \imp{-10\%} & \textbf{1728} \imp{-14\%} \\
        + DAPO w \our   & \textbf{62.6} \imp{+2.2} & 79.0 \drp{-0.2} & 653 \imp{-7\%} & 1217 \imp{-5\%} & \textbf{30.4} \imp{+2.3} & 47.8 \imp{+0.8} & 1105 \imp{-5\%} & 2354 \drp{+7\%} \\
        \bottomrule
    \end{tabular}}
    \label{tab:math-2}
\end{table}

We provide a comprehensive comparison between \our\ and \old\ in Table \ref{tab:countdown}, \ref{tab:math-1} and \ref{tab:math-2} for Countdown and Math tasks, reporting their performance and completion length on test datasets after the complete 500 steps of training. Observations are summarized as follows:

\paragraph{\our\ delivers consistent gains in correctness across various benchmarks on both GRPO and DAPO.}
Across all task-policy combinations, \our\ outperforms \old\ in final Pass@1 scores. On the Countdown dataset, \our\ improves accuracy by an average of 4.5\% under both GRPO and DAPO. On the Math dataset, gains are averagely 3.8\%. Notably, GRPO + \our\ achieves comparable or even higher performance than DAPO + \old\, despite DAPO's computationally intensive mechanisms such as group filtering. Moreover, DAPO + \our\ achieves state-of-the-art performance on both benchmarks, reinforcing that trajectory diversity during rollout is the primary driver of performance gains. This finding aligns with ablation studies in DAPO \citep{dapo}, which identified rollout group filtering as the most effective component of their framework.

\paragraph{\our\ consistently reduces inference cost while enhancing performance.}
Beyond accuracy, \our\ significantly reduces the average length of generated reasoning trajectories at test time. On Countdown, completion length decreases by 22\% under both GRPO and DAPO; on math datasets, we observe a 8.3\% reduction. We attribute this dual benefit to \our's core mechanism: by encouraging exploration of diverse reasoning paths during training, it exposes the policy to a broader distribution of solutions, guiding the model to internalize efficient reasoning strategies. In contrast, \old\ tends to traverse the reasoning space sequentially within independent trajectories due to its insufficient parallel exploration, often resulting in verbose, redundant, or over-elaborated chains.

\subsection{Training Dynamics}
\label{sec:curve}

\begin{figure}
    \centering
    \begin{subfigure}{0.4\textwidth}
        \includegraphics[width=\textwidth]{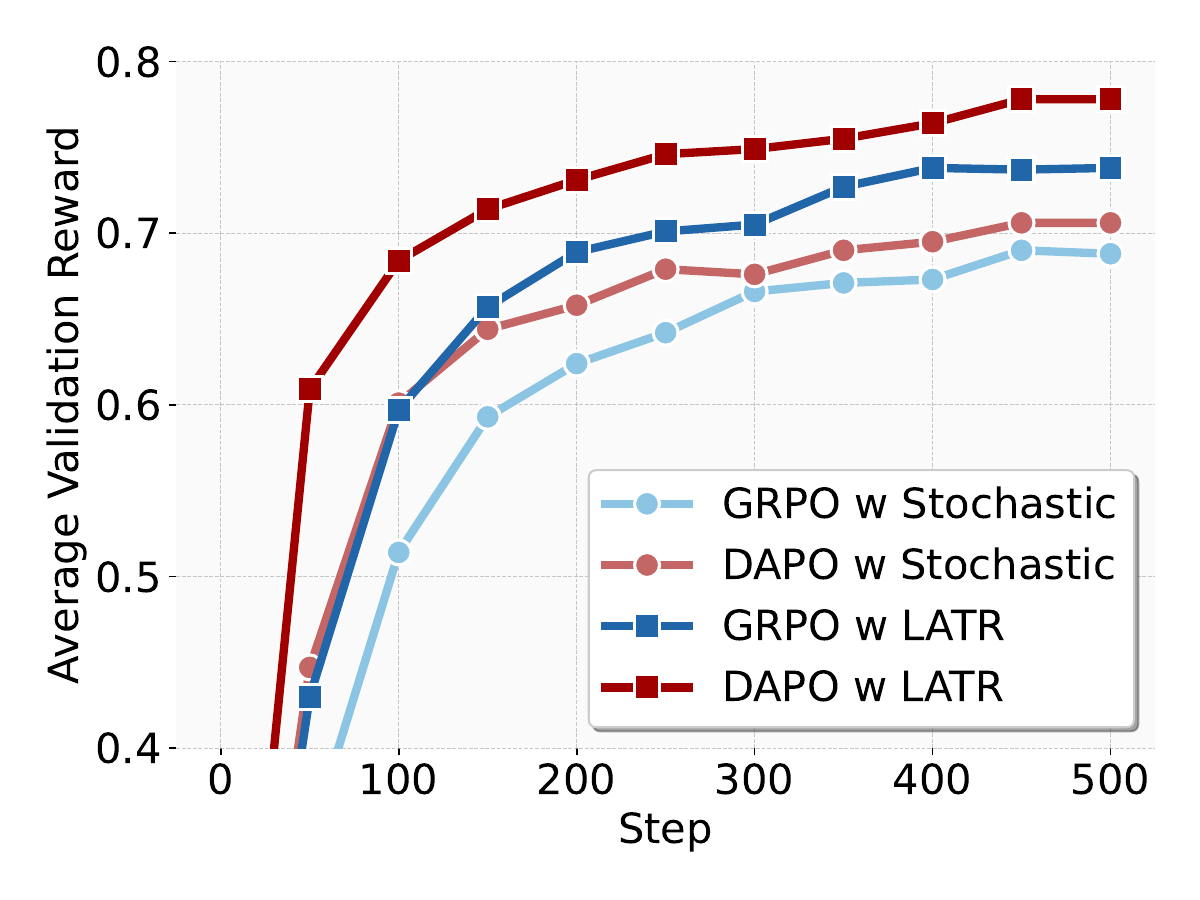}
    \end{subfigure}
    \hspace{2em}
    \begin{subfigure}{0.4\textwidth}
        \includegraphics[width=\textwidth]{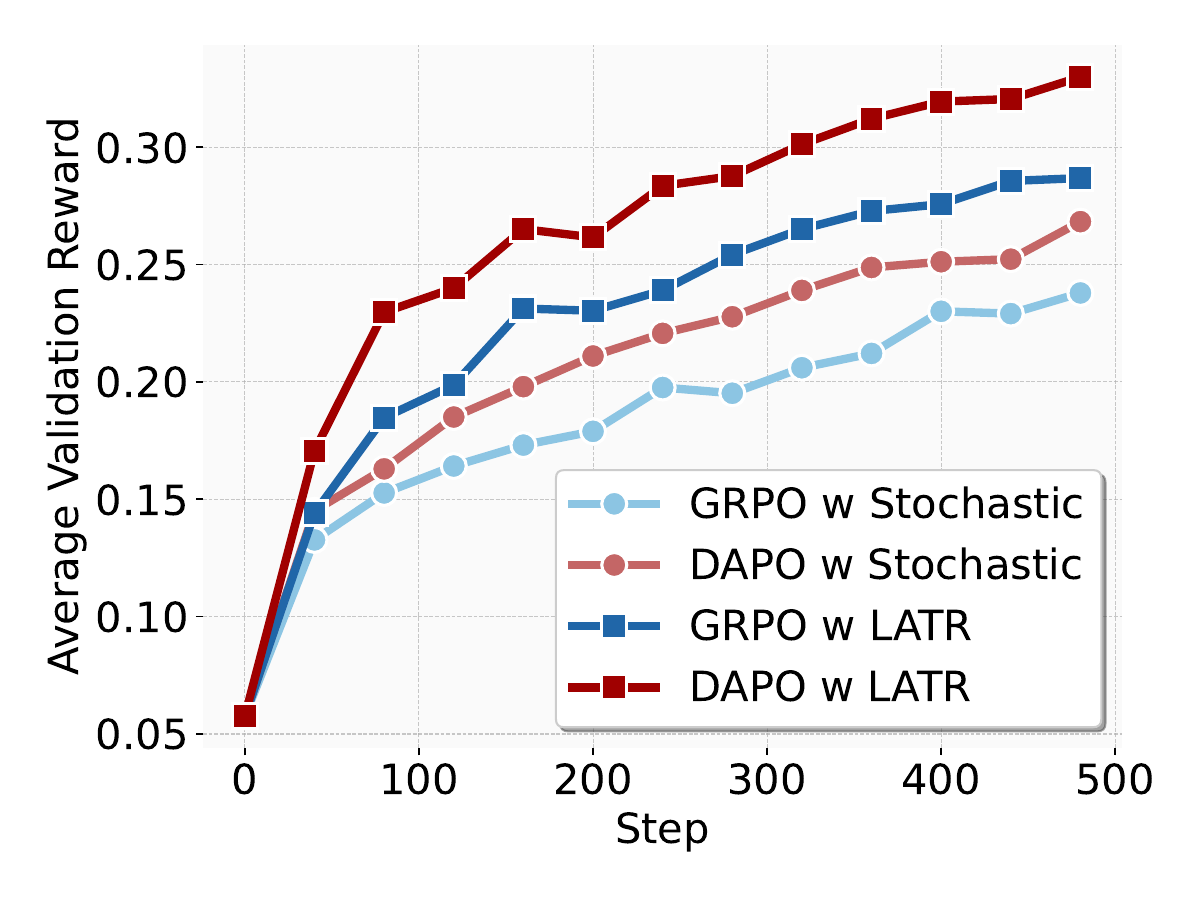}
    \end{subfigure}
    \vspace{-1em}
    \caption{Learning curve comparison on Countdown (left) and DAPO-Math (right) datasets.}
    \label{fig:curve}
    \vspace{-1em}
\end{figure}

To further investigate the RL training process with \our\ and \old, we analyze training dynamics by plotting validation accuracy against training step in Figure \ref{fig:curve}. The results reveal that \our\ not only converges to a better solution, but does so considerably faster.

Under DAPO, \old\ requires 450 steps to reach peak performance on the Countdown task, whereas \our\ achieves the same level of accuracy by step 150, resulting in a 3$\times$ acceleration in training efficiency. On the math task, compared to step 500 for \old, DAPO + \our\ reaches same performance at step 240, yielding a 2$\times$ speedup. Crucially, the acceleration provided by \our\ exceeds that gained by upgrading from GRPO to DAPO, despite DAPO's heavier data requirements and computational overhead per step. This suggests that \our's enhanced exploration of diverse trajectories is able to translate into more informative policy updates per training iteration. In effect, \our\ increases the sample efficiency of the RL process, enabling faster learning without architectural changes or additional data.

%% file: Sections/discussion.tex
\section{Discussions}
To evaluate the behavior and advantages of \our\ under varying conditions, we conduct a comprehensive set of controlled experiments. Unless otherwise specified, all analyses in this section are performed using the DAPO algorithm on the Countdown dataset, with all other hyperparameters and architectural settings held consistent with the main experiments. Comparison with other rollout strategies, impact of different similarity metrics for pruning, impact of branching and pruning thresholds, analyses on efficiency, and key statistics of \our\ are provided in Appendix \ref{app:analyses}.

\subsection{Diversity Comparison}
\label{sec:diverse}

To empirically validate that \our\ promotes greater diversity among reasoning trajectories within each rollout group, we conduct a comparative analysis between \our\ and the baseline method \old. We evaluate three variants of the Qwen2.5-3B architecture: Qwen2.5-3B, Qwen2.5-3B-Instruct, and Qwen2.5-3B trained with GRPO + \our, which we name it Qwen2.5-\our. This progression, from a raw pretrained model to an instruction-tuned variant and finally to a policy-optimized model incorporating \our, enables a nuanced assessment of how \our\ influences diversity across different stages of model development. In addition to standard performance metrics (Pass@1 and Pass@8), we also evaluate the average number of distinct final answers per rollout group. Two answer expressions of Countdown are considered distinct if their evaluated numerical outcomes differ. This ensures that diversity is measured in terms of semantic rather than syntactic variation.

\begin{wraptable}{r}{0.58\textwidth}
\vspace{-2em}
\hspace{0.4em}
\begin{minipage}{0.55\textwidth}
    \caption{Diversity comparison between \old\ and \our.}
    \adjustbox{width=\textwidth}{
    \begin{tabular}{lccc}
        \toprule
        Method & Pass@1 & Pass@8 & \# Ans. \\
        \midrule
        Qwen2.5-3B + Stoch.          & 5.8  & 28.9 & 6.3 \\
        Qwen2.5-3B + \our            & 6.1  & 30.7 & 6.9 \\
        \midrule
        Qwen2.5-3B-Instruct + Stoch. & 9.4  & 35.2 & 6.4 \\
        Qwen2.5-3B-Instruct + \our  & 10.9 & 40.6 & 6.9 \\
        \midrule
        Qwen2.5-3B-\our\ + Stoch.     & 70.9 & 77.4 & 2.6 \\
        Qwen2.5-3B-\our\ + \our       & 68.9 & 79.9 & 3.0 \\
        \bottomrule
    \end{tabular}}
    \label{tab:diverse}
\end{minipage}
\vspace{-2em}
\end{wraptable}

As shown in Table \ref{tab:diverse}, \our\ consistently yields higher Pass@8 scores and a greater number of distinct answers per rollout group across all three model variants compared to \old. These results support our claim that \our\ enhances intra-group trajectory diversity, thereby facilitating more effective policy learning through broader exploration of the solution space.

\subsection{Effect of Different Components}

\begin{wraptable}{r}{0.53\textwidth}
\vspace{-1em}
\hspace{0.4em}
\begin{minipage}{0.5\textwidth}
    \caption{Performance comparison on \old\ and variants of \our.}
    \adjustbox{width=\textwidth}{
    \begin{tabular}{lcccc}
        \toprule
        Method & Pass@1 & Pass@8\\
        \midrule
        Stochastic & 70.7 &	78.0 \\
        \our\ w rand branch & 69.6 &	75.8 \\
        \our\ w rand prune & 72.5 &	79.2 \\
        \our\ w/o prune & 71.0 &	78.7 \\
        \modify{\our\ w token-level lookahead} & 72.1 &	80.4 \\
        \our & \textbf{74.7} &	\textbf{81.5} \\
        \bottomrule
    \end{tabular}}
    \label{tab:ablation}
\end{minipage}
\vspace{-1em}
\end{wraptable}

We dissect the contributions of \our's core components through an ablation study. Specifically, we evaluate four variants of \our: (1) random branching, (2) random pruning, (3) no pruning, and \modify{(4) token-level lookahead in place of trajectory-level lookahead.} In the random variants, the branching or pruning ratio is matched to the average ratio observed in the full \our\ throughout training. In the token-level lookahead variant, pruning decisions are made solely based on the next token: a branch is pruned if the next tokens across trajectories are identical. This design enables us to isolate the effects of structured branching and similarity-based pruning on overall performance. Our findings are summarized below.

\paragraph{Random branching leads to unstable training and degrades final performance.}
As shown in Table \ref{tab:ablation}, \our\ with random branching performs even worse than \old. We observe that the KL divergence between the policy model and the base (reference) policy rises to as high as 1.0 within just 50 training steps, signaling severe off-policy behavior. This instability stems from uncontrolled branching. Without the probability thresholds imposed by our method, the model may generate extremely low-probability sequences that diverge significantly from the base policy, thereby disrupting the learning process.

\paragraph{Both random and no pruning yield suboptimal results.}
The variants of \our\ without pruning and with random pruning achieve only modest improvements over \old, confirming that branching alone enhances exploration by diversifying rollout trajectories. However, the full \our\ outperforms both. This performance gap is primarily attributable to trajectory-level redundancy, as rollout groups generated without pruning or with random pruning frequently contain sequences that follow similar reasoning paths, reducing effective diversity and leading to inefficient policy updates. Moreover, the comparison between the no-pruning and random-pruning variants highlights budget exhaustion as another critical factor. Without pruning, the fixed rollout budget $k$ is quickly depleted in early generation steps, leaving insufficient capacity for exploration in later stages.

\paragraph{Token-level lookahead underperforms trajectory-level lookahead.}
Although token-level lookahead outperforms both \old\ and the no-pruning variant, it falls significantly short of the full \our\ model. This deficit stems from its limited ability to capture trajectory divergence. Pruning decisions based solely on the next token are often inaccurate, leading to the premature removal of potentially valuable branches and degrading rollout quality.

In summary, while branching provides a robust mechanism for exploration, dynamic, similarity-aware pruning serves as a crucial regulator: it ensures that the exploration budget is allocated meaningfully across the generation process and effectively mitigates redundant trajectories.

\subsection{Scalability with Rollout Number}

The rollout budget $k$ fundamentally constrains the scope of exploration in RLVR training. We evaluate how \our\ and \old\ scale with increasing $k \in \{4,8,12,16\}$. Results in Figure \ref{fig:kt} reveal two critical trends: 

\begin{compactenum}[1)]
    \item \our\ consistently outperforms \old\ at every value of $k$, demonstrating robustness to budget constraints.
    \item While \old\ performance plateaus at $k=8$, \our\ continues to improve up to $k=12$, indicating a higher effective capacity for leveraging additional rollouts.
\end{compactenum} 

% \begin{wrapfigure}{r}{0.48\textwidth}
% \hspace{0.4em}
% \begin{minipage}{0.45\textwidth}
%     \includegraphics[width=\textwidth]{images/k.pdf}
%     \caption{Comparison of test correctness with different rollout number $k$ (\%).}
%     \label{fig:k}
% \end{minipage}
% \vspace{-3em}
% \end{wrapfigure}

This suggests that \our\ not only uses its budget more efficiently but also raises the performance ceiling of the system, enabling gains from larger $k$ values that \old\ cannot exploit.

\subsection{Impact of Different Sampling Temperatures}

% \begin{wrapfigure}{r}{0.48\textwidth}
% \vspace{-1em}
% \hspace{0.4em}
% \begin{minipage}{0.45\textwidth}
%     \includegraphics[width=\textwidth]{images/temp.pdf}
%     \caption{Comparison of test correctness with different temperature $t$ (\%).}
%     \label{fig:temp}
% \end{minipage}
% \vspace{-1em}
% \end{wrapfigure}

In standard RL frameworks with \old, the sampling temperature $t$ governs the exploration-exploitation trade-off: higher $t$ increases stochasticity and thus exploration, but risks degrading rollout quality. In contrast, \our\ delegates exploration primarily to its branching-and-pruning mechanism, using $t$ only to modulate stochastic fallback and hybrid rollouts, which is described in Section \ref{sec:optim}. 

\begin{figure}
    \centering
    \begin{subfigure}{0.4\textwidth}
        \includegraphics[width=\textwidth]{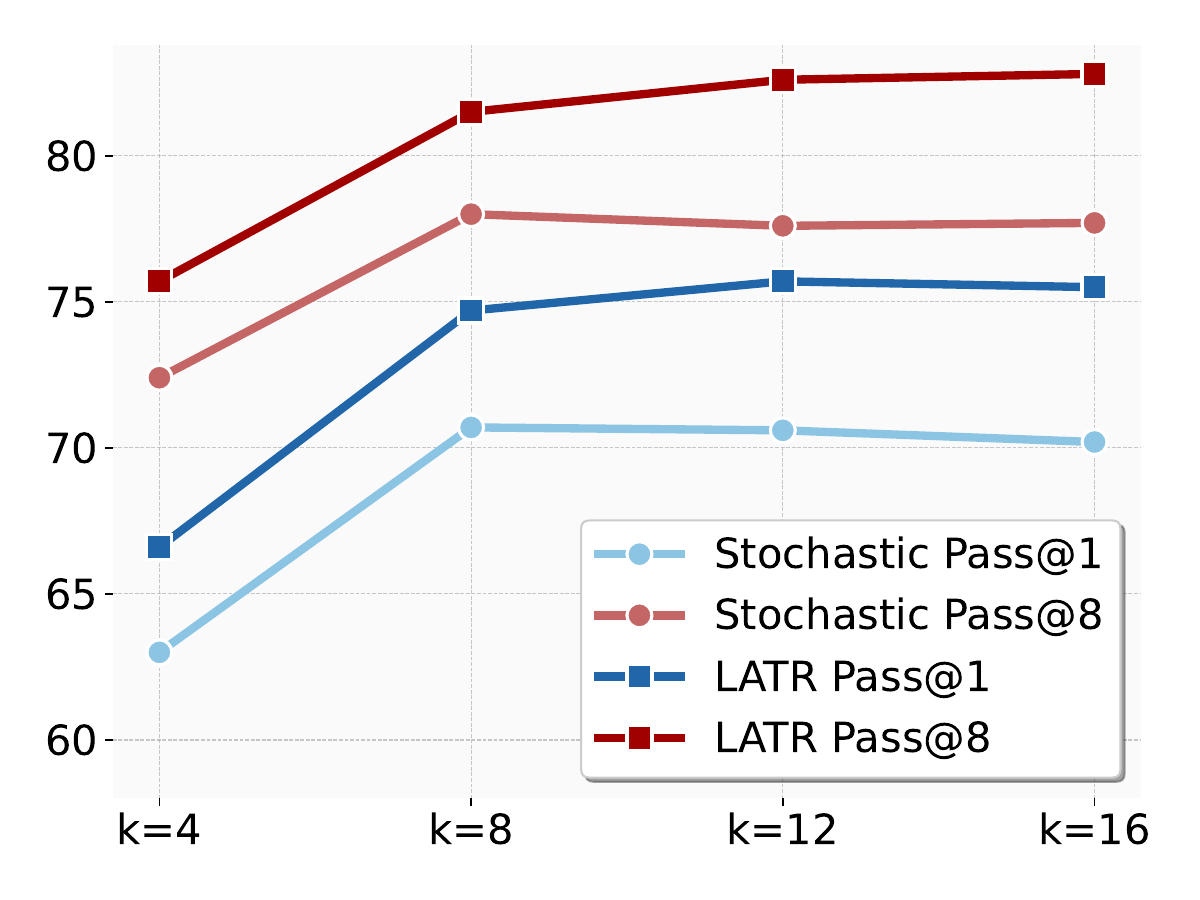}
    \end{subfigure}
    \hspace{2em}
    \begin{subfigure}{0.4\textwidth}
        \includegraphics[width=\textwidth]{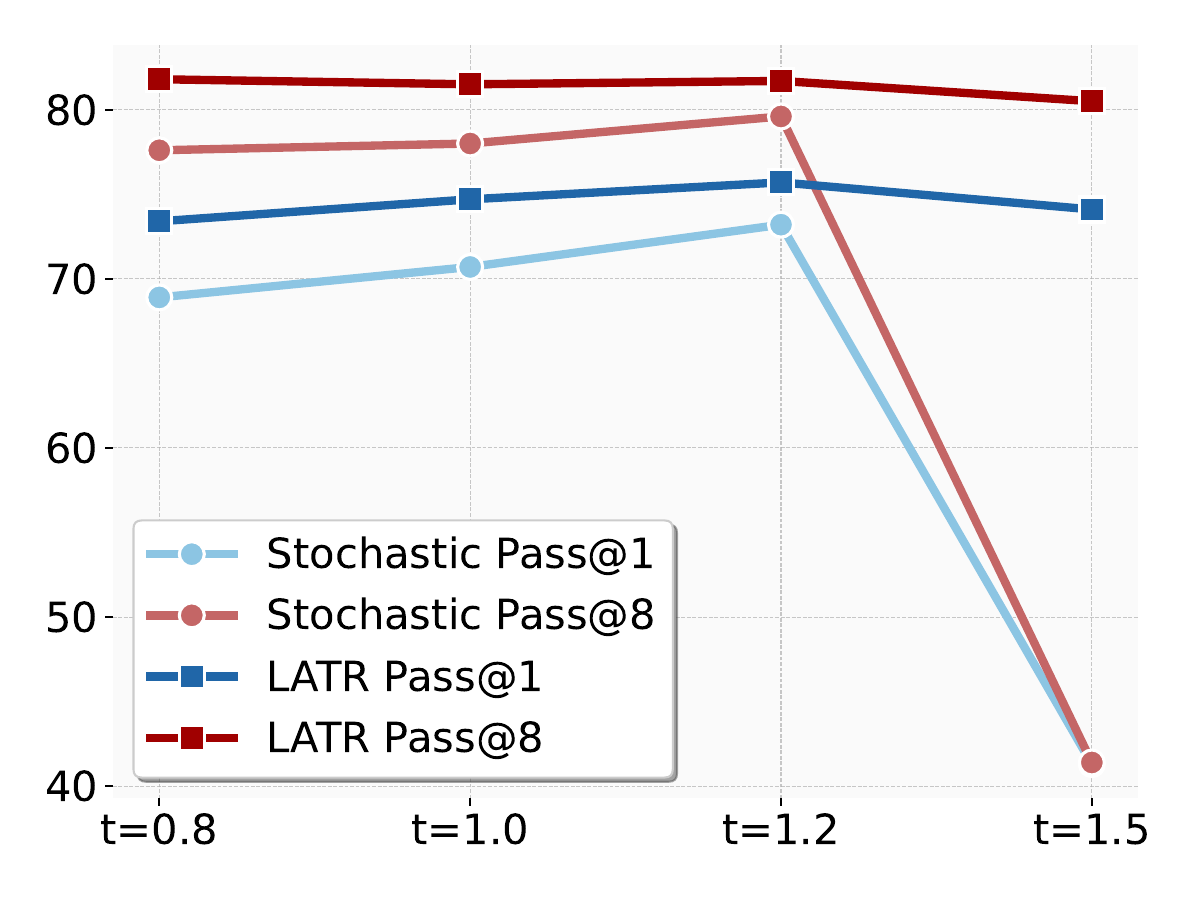}
    \end{subfigure}
    \caption{Comparison of test correctness with different rollout number $k$ and temperature $t$ (\%).}
    \label{fig:kt}
    \vspace{-1em}
\end{figure}

We evaluate performance across $t\in\{0.8,1.0,1.2,1.5\}$. As shown in Figure \ref{fig:kt}, both methods peak near $t=1.2$, suggesting this is optimal for the base policy. Notably, \our\ achieves superior performance at every $t$, and exhibits lower variance across temperatures. 

This robustness stems from \our's architectural decoupling: exploration is driven by structural diversity (branching + pruning), not sampling noise. Consequently, \our\ is less sensitive to suboptimal temperature tuning.

\begin{table}[b]
    \centering
    \vspace{-1em}
    \caption{Performance comparison of \our\ and \old\ on different base models.}
    \adjustbox{width=0.7\textwidth}{
    \begin{tabular}{lcccc}
        \toprule
        \multirow{2}*{Method} & \multicolumn{2}{c}{Correctness (\%) $\uparrow$} & \multicolumn{2}{c}{Average Length $\downarrow$} \\
        & Pass@1 & Pass@8 & Pass@1 & Pass@8 \\
        \midrule
        Qwen2.5-7B        & 1.1  & 5.4  & 556 & 997 \\
        Qwen2.5-7B + Stochastic & 70.9 & 79.8 & 522 & 722 \\
        Qwen2.5-7B + \our & \textbf{76.0 }& \textbf{82.1} & \textbf{396} & \textbf{508} \\
        \midrule
        Qwen3-1.7B-Base        & 2.0  & 9.5  & 542 & 916 \\
        Qwen3-1.7B-Base + Stochastic & 66.0 & 75.5 & 521 & 673 \\
        Qwen3-1.7B-Base + \our & \textbf{67.8} & \textbf{77.6} & \textbf{494} & \textbf{662} \\
        \bottomrule
    \end{tabular}}
    \label{tab:models}
\end{table}

\subsection{Generalizability Across Different Base Models}

To evaluate the generalizability of \our\ across diverse base models, we conduct additional experiments using additional models from different series and scales, specifically Qwen2.5-7B and Qwen3-1.7B-Base. As shown in Table \ref{tab:models}, \our\ consistently outperforms \old\ across all evaluated models, demonstrating the broad applicability and robustness of our proposed method.

%% file: Sections/related_work.tex
\section{Related Work}

\subsection{Reinforcement Learning with Verifiable Rewards}

Reinforcement Learning with Verifiable Rewards (RLVR) has emerged as a powerful alternative for tasks with verifiable results \citep{lambert2024tulu}. In RLVR, the reward signal is derived from an external verifier, providing an objective measure of a trajectory's success. Within RLVR, GRPO \citep{deepseekmath} have become the state-of-the-art solution. Rather than relying on a learned value model, it compares trajectories within a sampling group and updates policy based on relative success.

Following this line of research, many seek to improve the performance of GRPO. DAPO introduces clip higher technique and removes RL constraints to enable aggressive policy updates towards correct reasoning. GSPO \citep{gspo} proposes sequence-level rewards to smooth and stabilize learning. These innovations on policy update are orthogonal to the rollout strategy, so \our\ is fully compatible with these methods. Replacing vanilla rollouts with \our-generated trajectories yields additive performance improvements, as we demonstrate empirically.

A few works have also touched upon the rollout strategy, though typically as a secondary component. DAPO \citep{dapo} proposes a group filtering strategy to oversample and discard groups with identical rewards. ProRL \citep{prorl} increases the sampling temperature to obtain more diverse rollout sequences. Despite these advancements, these methods only address trajectory-level in-group diversity indirectly. Their reliance on token-level stochastic sampling is prone to generating semantically redundant reasoning paths, a limitation our work directly confronts.

More recently, two contemporaneous works integrate tree search into RLVR. TreeRL \citep{treerl} and TreePO \citep{treepo} propagate sparse binary outcome rewards backward through the reasoning tree to derive dense process rewards that guide policy updates. TreePO additionally enhances generation efficiency by reusing shared prefixes and pruning unpromising branches early in the rollout process. While both methods leverage tree-based structures, their objectives differ fundamentally from ours, as they primarily aim to refine reward estimation or improve computational efficiency. In contrast, we adopt tree-search to explicitly foster and compare diverse reasoning trajectories within a rollout group. This trajectory-level diversity enriches the reward signal by capturing a broader spectrum of potential outcomes, thereby enhancing policy learning.

\subsection{Lookahead Reasoning for LLMs}
Recent work has increasingly explored lookahead-based reasoning strategies in LLMs, with their majorly focus on inference-time approaches or offline data construction. Tree-of-Thoughts (ToT) \citep{tot} pioneered this direction by generating multiple reasoning branches at each step and selecting the most promising path using an external reward model. Subsequent methods such as MCTS-DPO \citep{mcts-dpo} and ReST-MCTS \citep{rest-mcts} extend this idea by integrating Monte Carlo tree search with lookahead estimation to decompose sparse, instance-level rewards into dense, step-level supervision signals. Quiet-STaR \citep{quiet-star} also leverages a lookahead mechanism, generating token-wise rationales that anticipate future text and optimizing them based on their contribution to correct continuations. 

While these works share the common ingredient of lookahead search, their objectives differ fundamentally from ours. Our primary goal in employing lookahead tree search is not reward propagation or step-level supervision, but rather to explicitly compare and promote trajectory-level diversity among rollouts for the same problem. This explicit focus on trajectory-level diversity distinguishes our method from prior lookahead approaches in LLMs.

%% file: Sections/appendix.tex
\appendix

\section{LLM Usage}
In the course of preparing this manuscript and supporting materials, we leveraged large language models (LLMs) as auxiliary tools to enhance the efficiency and quality of non-core research tasks. Specifically, LLMs were employed in two primary capacities: 

\begin{compactenum}[1)]
    \item \textbf{Language polishing}: We used LLMs to assist in proofreading, grammatical correction, and stylistic refinement of the manuscript's prose.
    \item \textbf{Boilerplate and utility code generation}: For ancillary implementation tasks, such as file I/O wrappers, format converters, or logging utilities, we used LLMs to accelerate prototyping.
\end{compactenum}

\section{Details on Experiment Setup}
\label{app:detail}
In this section, we detail the datasets, evaluation protocols, and implementation configurations.

\subsection{Datasets and Task Formulations}

\paragraph{Logical Reasoning.}
We adopt the Countdown dataset \citep{tinyzero}, which challenges models to construct arithmetic expressions from a given set of integers that evaluate exactly to a target number. Following \citet{tinyzero}, we define a two-component reward function:
\begin{equation}
R = 0.1 \cdot \mathbb{I}_{\text{format}} + 0.9 \cdot \mathbb{I}_{\text{correct}},
\end{equation}
where $\mathbb{I}_{\text{format}}$ indicates syntactic validity and $\mathbb{I}_{\text{correct}}$ indicates semantic correctness. Models are trained on the training split and evaluated on the official test set.

\paragraph{Mathematical Problem Solving.}
For mathematical reasoning, we train on the DAPO Math dataset \citep{dapo}, a curated collection of problems drawn from diverse sources. Consistent with \citet{dapo}, the reward is binary:
\begin{equation}
R = \mathbb{I}_{\text{correct}},
\end{equation}
awarding 1.0 only for exact numerical matches.

To ensure broad generalization, we evaluate not only on DAPO Math's held-out validation set, which is manually partitioned with 1,024 samples, but also on three established external benchmarks, including MATH500 \citep{math500}, AMC2023 \citep{amc}, and OlympiadBench \citep{olympiad}.

\begin{figure}
    \centering
    \includegraphics[width=0.95\linewidth]{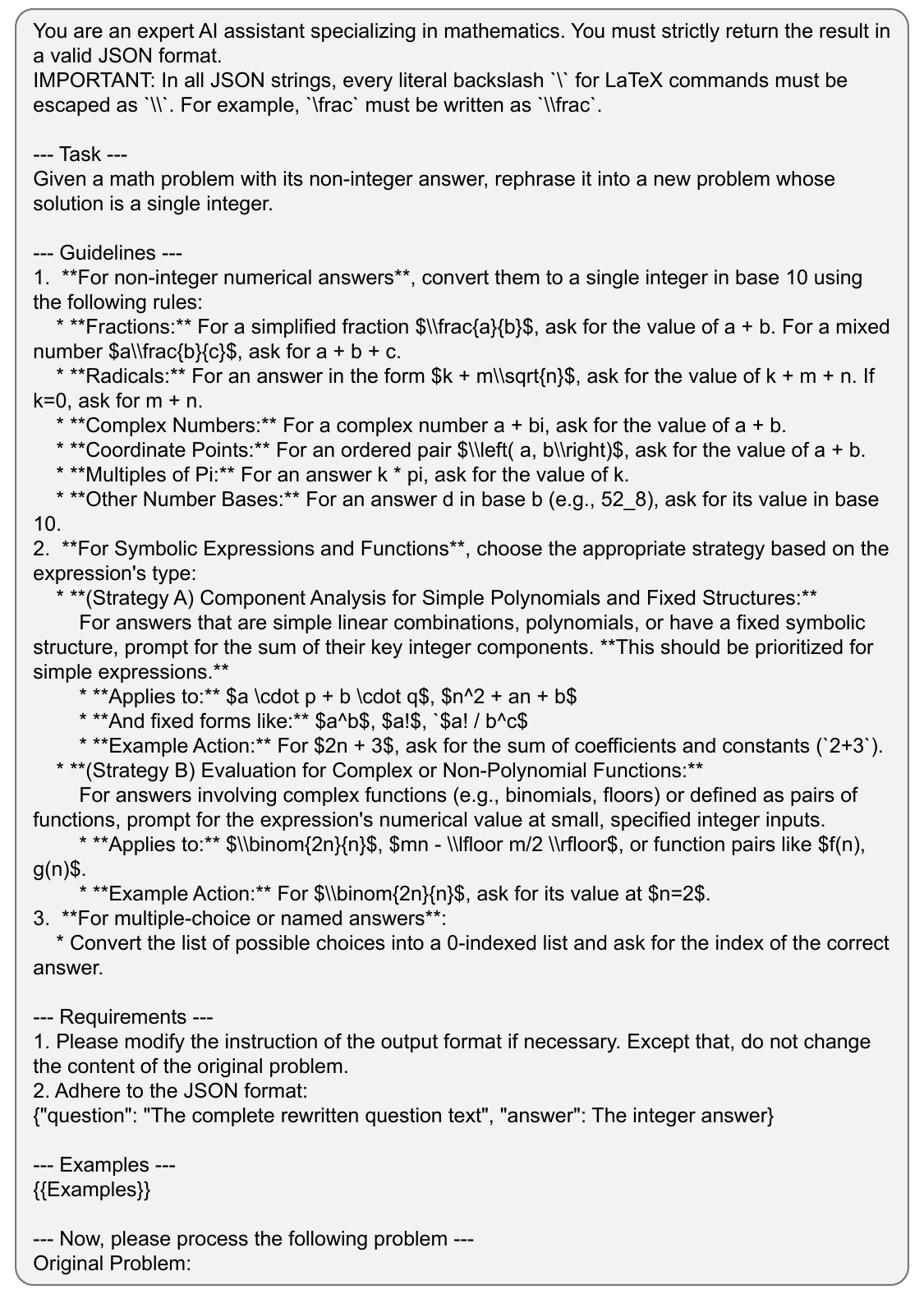}
    \caption{Prompt for data transformation.}
    \label{fig:prompt}
\end{figure}

To maintain consistency across datasets with heterogeneous answer formats, following \citet{dapo}, we apply a standardized answer normalization pipeline that maps all results to integers. We construct a comprehensive few-shot prompt that instructs Gemini-2.5-pro \citep{gemini} to apply a set of deterministic heuristics according the original answer's format. These heuristics include: (1) for structured non-integer answers like fractions ($p/q$) or radicals ($k + m\sqrt{n}$), rephrasing the question to ask for the sum of their components (e.g., $p+q$ or $k+m+n$); (2) for symbolic expressions, either summing the coefficients of simple polynomials or evaluating complex functions when assigning the variables (e.g., $x=2$); and (3) for multi-part or multiple-choice answers, asking for the sum of solutions or the 0-indexed position of the correct choice. The few shot prompt applied is provided in Figure \ref{fig:prompt}.

\subsection{Evaluation Protocol}

We perform stochastic sampling on the trained model for a fixed 8 times for each sample in the evaluation datasets, and report Pass@1 (the average accuracy over a single sampled completion per question) and Pass@8 (the accuracy of the best solution among 8 independently sampled completions per question). In addition to correctness, we also include average completion length for both Pass@1 and Pass@8 to quantify test-time computational cost and efficiency.

\subsection{Implementation Details}
\label{sec:detail}

\paragraph{Sampling and Rollout Parameters.}
We set our sampling parameters following \citet{dapo, tinyzero}. During training, we sample rollouts with temperature = 1.0, top-$k$ = $-1$, and top-$p$ = 1.0 to encourage exploration. During evaluation, we use temperature = 0.6, top-$k$ = 20, and top-$p$ = 0.95 for calibrated diversity. Each training step involves $k = 8$ rollouts per prompt. Maximum completion lengths are set to 1,024 tokens for Countdown and 8,192 tokens for math problems.

\paragraph{Algorithmic Parameters for \our.}
Hybrid rollout coefficient $\eta_0 = 1.0$, decaying per-step via $\eta_t = \eta_0 \cdot \gamma^t$, with $\gamma = 0.985$ (Countdown) and $\gamma = 0.995$ (Math). For branching thresholds, absolute probability threshold $\tau_{\text{abs}} = 0.25$, relative probability threshold $\tau_{\text{rel}} = 0.15$, and edit-distance threshold $\tau_{\text{ed}} = 0.4$. Lookback step $r$ is $\{20,30,50\}$, which means we enforce conditions on all of the 3 lookback windows, and all should be satisfied for a branch to be kept.

\paragraph{Training Hyperparameters.}
For training parameters, global data batch size is 256, global mini batch size is 256, local micro batch size is 8 for Countdown and 4 for DAPO Math, clip ratio is 0.2, KL penalty $\beta$ is 0.01. For DAPO, clip ratio high is 0.28, low is 0.2, and oversampled data generation batch size is 384. We train the Qwen2.5-3B base model on both datasets for a fixed 500 steps with AdamW optimizer and a constant learning rate of 1e-6. All our experiments are performed with VeRL-0.5.0 framework \citep{verl} on 8$\times$ NVIDIA H200 GPUs with mixed precision.

\section{Additional Analyses}
\label{app:analyses}

\subsection{Comparison with Alternative Rollout Strategies}

To further demonstrate the effectiveness of \our, we compare it against two other baseline rollout strategies:

\begin{compactenum}[1)]
    \item \textbf{Rollout Down-sampling (RDS)} \citep{pods}: Similar to the group filtering in DAPO, RDS also seeks to enhance trajectory diversity in a post-hoc manner. Specifically, it first generates $k=16$ trajectories via standard rollout and then selects the 8 most diverse trajectories for policy updates. The selection is implemented greedy, using the average of sentence-level BLEU and ROUGE scores as trajectory similarity measures.
    \item \textbf{Entropy Guided Tree Search (EPTree)} \citep{treerl}: Proposed in TreeRL, EPTree constructs a rollout tree to support fine-grained reward estimation and optimization during policy updates. After generating $M$ complete sequences, it identifies the top-$N$ tokens with the highest entropy and re-generates continuations $T$ times from each of these tokens, yielding a total of $M \times (N \times T + 1)$ sequences. Following the setup in TreeRL, we use $(M, N, T) = (4, 2, 1)$, resulting in 10 sequences per rollout. To ensure a fair comparison, we randomly sample 8 trajectories from these 10 for policy updates. We directly use the official code from TreeRL and integrates EPTree rollout into the VeRL training framework.
\end{compactenum}
% : initially, the two most dissimilar trajectories are chosen; subsequently, each new trajectory is selected to maximize its minimum distance to all previously selected ones. This process continues iteratively until 8 trajectories are retained. Trajectory similarity in SR is measured using the average of sentence-level BLEU and ROUGE scores.

\begin{wraptable}{r}{0.6\textwidth}
% \vspace{-1em}
\hspace{0.4em}
\begin{minipage}{0.57\textwidth}
    \caption{Performance comparison of different rollout strategies on the Countdown dataset.}
    \adjustbox{width=\textwidth}{
    \begin{tabular}{lcccc}
        \toprule
        \multirow{2}*{Method} & \multicolumn{2}{c}{Correctness (\%) $\uparrow$} & \multicolumn{2}{c}{Average Length $\downarrow$} \\
        & Pass@1 & Pass@8 & Pass@1 & Pass@8 \\
        \midrule
        GRPO w Stoch. & 65.9 & 73.9 & 473 & 610 \\
        GRPO w RDS    & 68.7 & 75.7 & \textbf{365 }& 489 \\
        GRPO w EPTree & 65.3 & 73.5 & 471 & 599 \\
        GRPO w \our   & \textbf{70.9} & \textbf{77.4} & 378 & \textbf{469} \\
        \midrule
        DAPO w Stoch. & 70.7 & 78.0 & 483 & 630 \\
        DAPO w RDS    & 68.5 & 74.4 & \textbf{348} & 462 \\
        DAPO w EPTree & 66.3 & 74.6 & 450 & 595 \\
        DAPO w \our   & \textbf{74.7} & \textbf{81.5} & 367 & \textbf{453} \\
        \bottomrule
    \end{tabular}}
    \label{tab:others}
\end{minipage}
% \vspace{-1em}
\end{wraptable}

As shown in Table \ref{tab:others}, \our\ consistently outperforms both RDS and EPTree across both GRPO and DAPO policy update algorithms. Notably, while GRPO combined with RDS yields improvements over \old\ due to enhanced trajectory diversity, the same combination under DAPO fails to surpass \old's performance. Further analysis reveals that DAPO + RDS leads to unstable training dynamics, marked by performance degradation and sharp increases in KL divergence during later training stages. This instability likely stems from the diversity-oriented selection mechanism, which biases towards selecting low-probability trajectories, thereby increasing off-policy risk. When combined with DAPO, which already promotes diversity through group-based filtering, this effect is amplified, ultimately contributing to model collapse.

In contrast, the underwhelming performance of EPTree suggests that the gains reported in TreeRL primarily arise from its novel policy update mechanism rather than its rollout strategy. Specifically, TreeRL employs Monte Carlo Tree Search (MCTS) to estimate fine-grained rewards for individual tree branches by propagating sparse binary outcome rewards backward through the tree, enabling targeted optimization of intermediate reasoning steps. By contrast, \our\ improves RL performance by enhancing trajectory-level diversity without requiring modifications to the underlying policy update algorithm.

\subsection{Impact of Different Similarity Metrics}
\label{app:similarity}

As described in Equation \ref{eq:prune}, the main experiments employ edit distance as the similarity measure to identify and prune redundant trajectory branches. In principle, however, numerous alternative metrics could be used to assess the divergence between partial sequences. To investigate this, we evaluate three additional similarity measures in this section: 

\begin{compactenum}[1)]
    \item \textbf{ROUGE-L}: defined as the ratio of the length of the longest common subsequence to the sequence length.
    \item \textbf{Suffix matching}: defined as the ratio of the length of the longest suffix of one sequence that appears anywhere in the other sequence to the sequence length.
    \item \textbf{Embedding-based}: the cosine similarity of the sequence embeddings calculated by the model Qwen3-Embedding-0.6B.
\end{compactenum}

For each metric, we conduct experiments while tuning the pruning threshold to identify its optimal value. The best results, summarized in Table \ref{tab:similarity}, show that embedding-based similarity yields the weakest performance, while all other token-level metrics achieve comparable and better final accuracy. The failure of embedding-based similarity is likely to stem from the inability for embedding models to capture fine-grained logical distinctions, since these models are usually trained to discern topic-level differences. Therefore, Given its simplicity and competitive efficacy, we retain edit distance as our pruning criterion.

\subsection{Impact of Branching and Pruning Thresholds}

\begin{wraptable}{r}{0.4\textwidth}
\vspace{-1.2em}
\hspace{0.4em}
\begin{minipage}{0.36\textwidth}
    \caption{Performance comparison of different thresholds.}
    \adjustbox{width=\textwidth}{
    \begin{tabular}{lcc}
        \toprule
        Threshold & Pass@1 & Pass@8\\
        \midrule
        $\tau_{abs}=0.2$  & 71.4 & 77.2 \\
        $\tau_{abs}=0.25$ & 74.7 & 81.5 \\
        $\tau_{abs}=0.3$  & 72.3 & 78.5 \\
        \midrule
        $\tau_{rel}=0.1$  & 72.0 & 77.7 \\
        $\tau_{rel}=0.15$ & 74.7 & 81.5 \\
        $\tau_{rel}=0.2$  & 72.1 & 77.6 \\
        \midrule
        $\tau_{ed}=0.3$ & 73.8 & 80.5 \\
        $\tau_{ed}=0.4$ & 74.7 & 81.5 \\
        $\tau_{ed}=0.5$ & 74.1 & 81.4 \\
        $\tau_{ed}=0.6$ & 73.4 & 79.8 \\
        \bottomrule
    \end{tabular}}
    \label{tab:tau}
\end{minipage}
\vspace{-2em}
\end{wraptable}

% The three fundamental hyper-parameters in \our\ include the branching probability thresholds $\tau_{abs},\tau_{rel}$, the simulation step $r$, and the pruning edit distance threshold $\tau_{ed}$. In Section \ref{sec:temperature}, we studied the impact of different temperatures for \our. Since in those studies $\tau_{abs}$ and $\tau_{rel}$ are kept constant, the different temperatures changes the probability distribution, which is equivalent to changing $\tau_{abs}$ and $\tau_{rel}$ with fixed temperature.
To investigate the impact of the pruning threshold $\tau_{ed}$, we conduct a series of experiments with $\tau_{ed} \in \{0.3, 0.4, 0.5, 0.6\}$. The results are summarized in Table \ref{tab:tau}. We observe that $\tau_{ed}=0.4$ yields the best performance, while both lower and higher values lead to a decline in results. This behavior can be attributed to the trade-off imposed by the pruning threshold: excessively high values result in overly similar trajectories, which consume the rollout budget without promoting exploration, whereas excessively low values impose overly stringent constraints, leading to an insufficient number of viable trajectories.

For branching thresholds $\tau_{abs}, \tau_{rel}$, we experiment by fixing one threshold while adjusting the other one. As shown in Table \ref{tab:tau}, $(\tau_{abs}, \tau_{rel})=(0.25, 0.15)$ yields the best performance, corroborating the suitability of our selected hyperparameters.

\begin{table}[t]
    \centering
    \caption{Performance comparison of different similarity metrics.}
    \label{tab:similarity}
    \adjustbox{width=0.65\textwidth}{
    \begin{tabular}{lcccc}
        \toprule
        \multirow{2}{*}{Method} & \multicolumn{2}{c}{Correctness (\%) $\uparrow$} & \multicolumn{2}{c}{Average Length $\downarrow$} \\
        & Pass@1 & Pass@8 & Pass@1 & Pass@8 \\
        \midrule
        Edit Distance   & 74.7 & 81.5 & 367 & 453 \\
        ROUGE-L         & 73.9 & 80.5 & 390 & 486 \\
        Suffix Matching & 74.9 & 81.7 & 388 & 493 \\
        Embedding-based & 72.9 & 79.8 & 369 & 445 \\
        \bottomrule
    \end{tabular}}
% \vspace{-1em}
\end{table}

\subsection{Efficiency Analysis}

The search tree in \our\ is bounded by a maximum width corresponding to the rollout number $k$. Unlike \old, which perform forward passes on each sequence independently at each step, \our\ dynamically branches and prunes sequences, resulting in a sparser tree structure, particularly during early generation stages. Consequently, the actual number of FLOPs consumed by \our\ is strictly less than that of \old\ for the same settings.

\begin{wrapfigure}{r}{0.38\textwidth}
\vspace{-1em}
\hspace{0.4em}
\begin{minipage}{0.35\textwidth}
    \includegraphics[width=\textwidth]{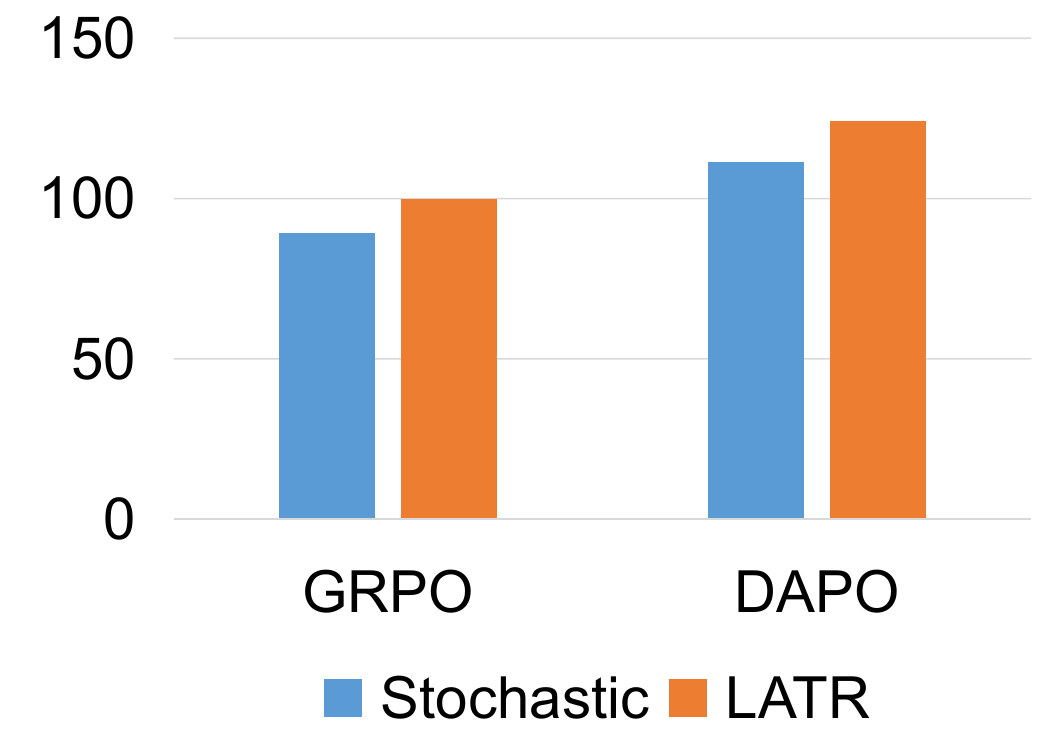}
    \caption{Comparison of average consumed time per training step under different settings (second).}
    \label{fig:eff}
\end{minipage}
\vspace{-2em}
\end{wrapfigure}

Empirically, \our\ exhibits a modest slowdown in generation speed during RL training compared to \old, as shown in Figure \ref{fig:eff}. Specifically, \our\ runs approximately 10\% slower per step than \old\ with the same configuration. However, compared with DAPO with \old, GRPO with \our\ is able to achieve comparable performance in shorter training time. Additionally, considering the averagely $2.3\times$ training speedup (as introduced in Section \ref{sec:curve}), \our\ is able to achieve higher performance in less total training time. This suggests that the algorithmic gains of \our\ outweigh its runtime penalties in end-to-end training scenarios.
% this efficiency cost is offset by the gain in task performance: \our\ with $k=8$ achieves results comparable to \old\ with $k=12$, which not only runs slower than \our\ but also demands substantially more VRAM. This trade-off highlights \our\ as a favorable approach for resource-constrained RL training, as it delivers high performance at lower memory cost and better FLOPs efficiency.

Further profiling reveals that the runtime overhead in \our\ primarily stems from the sequential computation patterns during the branching and pruning phase. Unlike \old, which processes contiguous batched inputs, \our\ dynamically inserts and removes sequences during tree expansion and pruning, so indexing and comparisons are performed per sequence rather than in a fully batched manner. Targeted optimizations, analogous to PagedAttention \citep{pagedattention} for \old, are likely to mitigate the overhead. While promising, such engineering improvements lie outside the scope of this work and are left to future efforts.

\subsection{Additional Statistics for \our}
\label{app:stat}

\begin{wraptable}{r}{0.63\textwidth}
\vspace{-1em}
\hspace{0.4em}
\begin{minipage}{0.6\textwidth}
    \caption{Key statistics for \our.}
    \adjustbox{width=\textwidth}{
    \begin{tabular}{lcc}
        \toprule
        Model & Branching Ratio & Saturation Length \\
        \midrule
        Qwen2.5-3B          & 0.101  & 65 \\
        Qwen2.5-3B-Instruct & 0.039  & 102 \\
        Qwen2.5-3B-\our     & 0.044  & 132 \\
        \bottomrule
    \end{tabular}}
    \label{tab:stat}
\end{minipage}
% \vspace{-2em}
\end{wraptable}

To further elucidate the behavior of our proposed method, we report two key statistics: the average branching ratio, which is the proportion of tokens at which new reasoning branches are initiated relative to the total number of generated tokens, and the average saturation length, defined as the number of tokens generated before early stopping is triggered. Following the setup in Section \ref{sec:diverse}, we present these metrics for the same three model variants: Qwen2.5-3B, Qwen2.5-3B-Instruct, and Qwen2.5-\our, enabling a consistent and comprehensive analysis across model stages.

As shown in Table \ref{tab:stat}, the branching ratios are consistently low across all models, indicating conservative branching behavior. Moreover, the average saturation length is notably shorter than the maximum completion length of 1,024 tokens. This observation aligns with prior findings \citep{early_token}, which suggest that the initial segments of a reasoning chain are often most critical in determining the final outcome.

